%% file: main.tex
\documentclass[11pt,letterpaper]{article}

\usepackage[left=1in,right=1in,top=1in,bottom=0.9in]{geometry}

\usepackage{graphicx}
\usepackage[table,xcdraw]{xcolor}
\usepackage{wrapfig,lipsum,booktabs}
\usepackage{setspace}
\usepackage{tikz}
\usepackage{booktabs}
\usepackage{tabularx}
\usepackage{longtable}
\usepackage{array}
\usetikzlibrary{decorations.shapes,shapes.geometric,shadows,arrows,automata,positioning,calendar,mindmap,backgrounds,scopes,chains,er,3d,matrix,fit}

\usepackage{wrapfig}

\usepackage[utf8]{inputenc}
\usepackage{amsmath}
\usepackage{amsfonts,times}
\usepackage{cite}

\usepackage{tikz,lipsum,lmodern}
\usepackage[most]{tcolorbox}

\usepackage{amssymb}
\usepackage{url}

\usepackage[compact]{titlesec}
    \titlespacing{\section}{0pt}{3ex}{2ex}
    \titlespacing{\subsection}{0pt}{2.0ex}{1.0ex}
    \titlespacing{\subsubsection}{0pt}{1.0ex}{0.5ex}

\usepackage{array}
\newcolumntype{L}{>{\centering\arraybackslash}m{3cm}}

\title{Insurance of Agentic AI}
\author{
Quanyan Zhu\thanks{Department of Electrical and Computer Engineering, New York University, Tandon School of Engineering. Email: \texttt{qz494@nyu.edu}. Office: Room 1004, 370 Jay Street, Brooklyn, NY 11201. Phone: 646.997.3371.}
}
\date{}

\begin{document}

\maketitle

\setstretch{1.67}
\pagenumbering{gobble}


\begin{abstract}
Agentic artificial intelligence (AI) systems are transforming the risk landscape by extending beyond information generation to autonomous planning, tool invocation, decision execution, and persistent modification of digital and physical environments. These capabilities introduce novel exposures that do not fit neatly within traditional insurance categories such as cyber, professional liability, product liability, or directors and officers coverage. This paper examines the emerging insurance market for agentic AI and develops a framework for understanding its underwriting, pricing, reinsurance, and product-design implications. We characterize agentic AI as a continuum of autonomy and delegated authority, emphasizing the distinction between informational outputs and systems capable of independently generating insured events through external actions. We analyze major risk pathways, including hallucinations, prompt-injection attacks, autonomous decision errors, model drift, dependency failures, and cyber-physical harms, and evaluate how existing insurance products are adapting to address these exposures. The paper further proposes an actuarial framework based on exposure assessment, scenario analysis, dependency mapping, and accumulation-risk management, drawing parallels to the evolution of cyber insurance. Finally, we present a coordinated insurance architecture that integrates cyber, technology errors and omissions, product liability, performance-warranty, and affirmative AI-liability coverages through explicit allocation mechanisms and dedicated AI aggregates. The analysis suggests that the future of agentic-AI insurance lies not in a single monoline product but in a layered ecosystem of complementary coverages supported by improved governance, transparency, telemetry, and regulatory clarity.
\end{abstract}

\section{Introduction}
\input{intro}

\section{Related Literature}
\input{literature}

 \section{Risk Landscape}

\input{risk}

 \section{Insurance Market}
\input{market}

 \section{Pricing and Reinsurance}

\input{pricing}

 \section{Product design}
\input{product}

\section{Conclusion}

Agentic AI represents a significant evolution in the nature of insurable technological risk. Unlike conventional software systems, agentic AI can plan, reason, invoke tools, coordinate actions across multiple systems, and produce persistent changes in digital and physical environments. As a result, insured losses may arise not only from cyber compromise but also from autonomous decision errors, hallucinations, prompt-injection attacks, model drift, unsafe delegation, dependency failures, and cyber-physical interactions. These characteristics create a risk profile that spans traditional insurance boundaries and challenges existing approaches to underwriting, claims allocation, and capital management.

The insurance industry has already begun adapting through a combination of expanded cyber policies, technology E\&O coverage, affirmative AI endorsements, standalone AI-liability products, performance-warranty structures, and coordinated multi-policy solutions. The evidence reviewed in this paper suggests that no single insurance product is likely to absorb the full spectrum of agentic-AI risk. Instead, the most sustainable market structure is likely to be a layered ecosystem in which different insurance mechanisms address distinct loss drivers while explicit allocation provisions manage mixed-cause events involving cyber incidents, AI behavior, operational failures, and physical damage.

From an actuarial perspective, the principal challenge is not merely limited historical loss experience but the combination of non-stationarity, dependency concentration, and evolving legal responsibilities. Consequently, underwriting and pricing must rely on exposure-based assessment, scenario analysis, accumulation-risk management, and control effectiveness rather than traditional credibility-based approaches alone. Lessons from the development of cyber insurance demonstrate that meaningful risk transfer can emerge before extensive claims histories are available, provided insurers maintain disciplined exposure measurement, robust governance standards, and appropriate reinsurance support.

Looking forward, the maturation of the agentic-AI insurance market will depend on several foundational developments: affirmative and transparent policy wording, AI-specific claims taxonomies, improved telemetry and auditability, consistent incident reporting, and clearer allocation of legal responsibility among developers, deployers, integrators, and end users. Regulatory initiatives such as the European Union AI Act and evolving industry standards are likely to contribute to greater predictability in both liability and claims outcomes. As empirical experience accumulates, insurers will be able to refine actuarial models, develop more sophisticated accumulation controls, and expand underwriting capacity. Ultimately, agentic AI should be viewed not merely as an emerging source of uncertainty but as an opportunity for insurance innovation. Organizations that successfully integrate governance, transparency, and risk-transfer mechanisms will be better positioned to realize the benefits of autonomous AI systems while maintaining resilience against their evolving operational, legal, and economic risks.

\bibliographystyle{abbrv}	
\bibliography{refs}

\end{document}

%% file: intro.tex
Agentic AI is emerging as a significant new source of insurable risk, although it has not yet crystallized into a standalone insurance class comparable to cyber, directors and officers (D\&O), or product liability coverage. Rather than constituting a single category of exposure, agentic AI introduces a spectrum of risks associated with systems capable of planning, invoking tools, executing multi-step tasks, and producing persistent changes in external environments. This conceptual boundary aligns with the National Institute of Standards and Technology (NIST)'s characterization of AI agents as systems capable of generating persistent external-state changes \cite{nistAgents}. It also aligns with game-theoretic accounts of agentic-AI cybersecurity, in which autonomous systems, attackers, defenders, and human overseers interact strategically across technical and organizational workflows \cite{zhu2025AgenticAIGameTheory}. From an insurance perspective, the relevant exposures extend well beyond conversational outputs to encompass actions such as authorizing transactions, sending communications, modifying software, updating organizational records, browsing autonomously on behalf of users, and interacting with physical or cyber-physical systems.

The insurance industry is responding through an increasingly sophisticated and layered risk-transfer architecture. Existing cyber, technology errors and omissions (E\&O), professional liability, media liability, general liability, product liability, and D\&O policies already address portions of the risk landscape. Concurrently, insurers are developing affirmative AI-specific endorsements and dedicated coverage solutions, exemplified by products such as AXA XL's Generative AI cyber endorsement \cite{axaGenAI} and emerging AI liability and performance-risk structures. As a result, the central market challenge is no longer whether agentic AI creates insurable exposures, but how insurance frameworks should allocate responsibility when losses arise from the intersection of AI behavior, cyber incidents, professional services, operational decisions, and product performance.

A particularly important underwriting consideration is that agentic-AI losses may occur independently of traditional cyber compromise. Significant losses can result from hallucinations, flawed reasoning, prompt-injection attacks, unsafe delegation chains, model drift, or unauthorized autonomous actions even in the absence of a network breach or security intrusion. Recent public evaluations by leading AI developers, including Anthropic and OpenAI, identify prompt injection, harmful task execution, and irreversible real-world actions among the principal operational risks associated with advanced AI agents. Anthropic's Claude 4 system card, for example, documented opportunistic blackmail behavior under a constrained shutdown scenario \cite{anthropicClaude4}. Although such evaluations do not constitute actuarial loss experience, they provide valuable evidence regarding the emerging risk mechanisms that insurers must assess beyond conventional network-security controls.

While the actuarial foundations of agentic-AI insurance remain under development, the industry's experience with cyber insurance suggests a viable path forward. Publicly available loss data specific to agentic AI remains limited, but cyber insurance itself matured through the use of exposure assessment, scenario analysis, catastrophe modeling, reinsurance support, and accumulation-risk management before extensive historical claims data became available \cite{naicCyber2025}. Agentic-AI underwriting is likely to evolve along a similar trajectory, incorporating scenario-based pricing, control-adjusted exposure measurement, vendor and model concentration analysis, and rigorous stress-testing methodologies. As operational experience accumulates, these approaches can provide the foundation for increasingly refined actuarial models.

The regulatory environment is also evolving in ways that may enhance insurability. Frameworks such as the European Union's AI Act establish governance, transparency, and accountability requirements for AI systems \cite{euAIAct2024}. Parallel regulatory initiatives, judicial developments, and enforcement actions are progressively clarifying the legal duties and liabilities associated with AI deployment. For insurers, greater legal clarity can improve risk assessment, claims forecasting, and defense-cost estimation, even in jurisdictions where AI-specific insurance remains voluntary rather than mandatory. In this sense, regulatory maturation may contribute not only to accountability but also to the development of more predictable insurance markets.

Agentic AI represents a promising new frontier for insurance innovation rather than merely a source of emerging uncertainty. Insurers can strengthen market resilience by replacing silent AI exposure with affirmative coverage language, underwriting according to levels of autonomy and action authority, establishing dedicated AI aggregates and mixed-causation allocation provisions, and leveraging quota-share and aggregate reinsurance structures while empirical loss experience remains limited. Policymakers, meanwhile, can support sustainable market development through initiatives that promote transparency, auditability, incident reporting, and clear liability allocation. The emergence of affirmative AI-liability products and specialized risk-transfer mechanisms indicates that this evolution is already underway, laying the groundwork for a mature and scalable market for agentic-AI risk \cite{armillaLloyds}.

%% file: literature.tex
The paper builds on several connected strands of literature. The first concerns AI safety and foundation-model risk. Amodei et al. identify accident-risk mechanisms such as side effects, reward hacking, distributional shift, scalable supervision, and unsafe exploration, all of which become more consequential when an AI system can act through external tools rather than merely generate advice \cite{amodei2016ConcreteAISafety}. The foundation-model literature further emphasizes that broadly trained models create both cross-domain opportunity and cross-domain risk because the same technical system can be adapted across many downstream contexts \cite{bommasani2021FoundationModels}. This observation is directly relevant to insurance, since model reuse and vendor concentration can turn what appears to be a local deployment risk into a correlated portfolio exposure. The broader literature on large language model risks also highlights dataset quality, opacity, documentation, bias, environmental costs, and social harms as governance problems that should be addressed before deployment at scale \cite{bender2021StochasticParrots}. More recent work on extreme AI risk argues for rigorous assessment, auditability, and adaptive governance as frontier systems become more capable and autonomous \cite{bengio2024ExtremeAIRisks}. These studies motivate the paper's emphasis on underwriting autonomy, delegated authority, monitoring, and post-loss evidence rather than treating AI as a generic software exposure.

A second strand concerns adversarial machine learning and robustness. Early work on adversarial examples showed that small, carefully chosen perturbations can cause neural networks to misclassify inputs in ways that transfer across models \cite{szegedy2013IntriguingProperties}. Goodfellow et al. provided a widely cited explanation and training method for adversarial examples, demonstrating that model fragility can arise even when inputs appear benign to human observers \cite{goodfellow2015AdversarialExamples}. Carlini and Wagner later showed why robustness claims require careful evaluation and why proposed defenses may fail under stronger attacks \cite{carliniWagner2017Robustness}. For agentic AI, these insights extend beyond image classifiers or static prediction systems. Prompt injection, jailbreaks, malicious tool outputs, and manipulated context can be understood as modern forms of adversarial input that affect not only classification but also downstream action. This is why the insurance problem cannot be reduced to model accuracy; it must account for adversarial interaction, control failure, and the consequences of external execution.

A third strand is the economics of cybersecurity, cyber insurance, and incentive design. The Gordon--Loeb model established a foundational economic framework for determining information-security investment under risk \cite{gordonLoeb2002InfoSecurityInvestment}. Subsequent cyber-insurance research examined insurance not merely as loss transfer but as a mechanism that can influence security incentives, especially when risks are correlated or interdependent \cite{bolotLelarge2009CyberInsurance}. Empirical and policy-oriented work on cyber-insurance markets has also shown that actual policy wording, coverage exclusions, pricing factors, and claims categories matter for whether insurance improves risk management or simply transfers poorly understood losses \cite{romanosky2019ContentAnalysisCyberInsurance}. These themes are central to agentic-AI insurance because insureds, model vendors, attackers, and insurers interact strategically under asymmetric information. The literature therefore supports a contract-design approach in which coverage, pricing, monitoring, and incentives are coordinated rather than handled as separate underwriting questions.

Within this cyber-insurance literature, several papers by Zhu and coauthors are especially relevant to the present article. Zhang and Zhu develop attack-aware cyber-insurance models for interdependent computer networks, showing how the user, attacker, and insurer can be modeled jointly through a bilevel game that combines adversarial behavior with moral hazard and incentive-compatible insurance design \cite{zhangZhu2016AttackAwareCyberInsurance}. Their later work on optimal cyber-insurance contract design extends this view to dynamic risk management, using principal-agent models and Markov decision processes to capture evolving cyber risks, user self-protection decisions, and risk-compensation effects \cite{zhangZhu2021OptimalCyberInsurance}. The FlipIn framework further applies game-theoretic cyber-insurance design to Internet-of-Things networks, where interdependent devices, advanced persistent threats, and network influence effects shape both insurability and welfare-maximizing coverage \cite{zhangZhu2019FlipIn}. Complementing this insurance-contract literature, Chen, Zhu, and Ba{\c{s}}ar study dynamic contract design for systemic cyber-risk management in interdependent enterprise networks, emphasizing asymmetric information between asset owners and cybersecurity managers \cite{chenZhuBasar2021DynamicContract}. Liu and Zhu's work on cyber-insurance risk-preference design and cyber insurance for cyber resilience adds a further layer by treating insured behavior, perception, monitoring, residual risk, and moral-hazard intensity as design variables that can improve cyber resilience rather than simply transfer losses \cite{liuZhu2022CyberMoralHazard,liuZhu2023CyberInsuranceResilience}. These papers provide a direct theoretical foundation for treating agentic-AI insurance as an incentive-compatible cyber-risk and resilience problem rather than as a passive extension of cyber liability coverage.

Agentic-AI insurance should be viewed as an interdisciplinary market-design problem. AI-safety research clarifies the technical failure modes, adversarial-ML research explains why model behavior can be manipulated or brittle under strategic inputs, and cyber-insurance economics explains why risk transfer must be coupled with incentives, controls, and credible evidence. This paper contributes by translating those insights into an insurance architecture for agentic AI: a framework for classifying autonomy, mapping loss pathways, pricing sparse and correlated risk, allocating mixed-cause claims, and designing affirmative coverage language that can mature as empirical claims experience develops.

%% file: risk.tex
Despite the rapid advancement and deployment of increasingly autonomous artificial intelligence systems, there remains no universally accepted legal, regulatory, or technical definition of \emph{agentic AI}. From an insurance and risk-management perspective, however, a functional definition is more useful than a purely semantic one. An AI system becomes meaningfully agentic when it extends beyond the generation of information and acquires the capacity to pursue objectives through multi-step reasoning, invoke external tools, maintain relevant state or memory, and produce persistent changes in external environments. The Organisation for Economic Co-operation and Development (OECD) adopts a broad and technology-neutral definition of AI systems \cite{oecdAIDefinition}, whereas the National Institute of Standards and Technology (NIST) adopts a narrower perspective that emphasizes an agent's ability to generate persistent external-state changes \cite{nistAgents}. Recent work on game theory and agentic AI reaches a similar conclusion from a security perspective: autonomy changes the risk problem because AI agents do not merely generate outputs, but participate in strategic workflows involving defenders, adversaries, users, and institutional controls \cite{zhu2025AgenticAIGameTheory}. For underwriting purposes, this distinction is particularly important because it separates \emph{information risk}, which arises from the generation of outputs and recommendations, from \emph{action risk}, which arises when AI systems are capable of autonomously executing decisions that may directly create insured losses.

Consequently, a practical insurance taxonomy of AI systems is more informative than debates over terminology. Agentic capability is best understood as a continuum characterized by increasing degrees of autonomy, delegation, and authority to affect external systems. At the lowest level are \emph{assistive systems}, such as conversational chatbots and coding copilots, which generate recommendations, text, or software artifacts but lack the authority to act beyond the immediate interaction. The next level comprises \emph{tool-enabled copilots}, in which AI systems may invoke application programming interfaces (APIs), retrieve information, or prepare actions for execution while material decisions remain subject to human approval. More advanced are \emph{autonomous digital agents}, which can browse websites, execute transactions, modify files, communicate with external parties, or interact directly with enterprise information systems. Beyond these are \emph{multi-agent systems}, in which multiple specialized agents coordinate activities, delegate subtasks, and collectively pursue higher-level objectives. At the highest level are \emph{cyber-physical agents}, whose actions extend into the physical domain through robotics, industrial control systems, autonomous vehicles, intelligent buildings, healthcare devices, Internet-of-Things (IoT) infrastructures, and energy-management systems. OpenAI's Operator platform provides a contemporary example of an intermediate-level agentic system because it combines planning capabilities, browser interaction, tool invocation, and autonomous execution of external actions \cite{openaiOperatorWeb}.

The distinction among these categories is not merely conceptual; it has direct implications for coverage determination, underwriting, and risk classification. A retrieval-augmented chatbot that provides information but lacks authority to alter records or initiate transactions may generate media-liability, privacy, professional-liability, or reputational exposures, yet it is not necessarily the primary target of agentic-AI coverage. Conversely, deterministic robotic process automation (RPA) systems may create significant operational risk without satisfying all contemporary definitions of artificial intelligence. Similarly, predictive models employed in underwriting, hiring, lending, or healthcare workflows clearly constitute AI systems, but they are not inherently agentic unless they independently initiate, authorize, or control downstream actions. NIST's emphasis on persistent external-state modification therefore provides a particularly useful underwriting criterion because it identifies systems capable of directly generating insured events rather than merely influencing human decision making \cite{nistAgents}.

From a risk-assessment perspective, the central underwriting question is no longer whether an organization uses artificial intelligence, but rather what authority has been delegated to the AI system, what controls govern its operation, what approvals are required prior to execution, and which external systems, assets, or processes it is permitted to modify. The degree of autonomy therefore becomes a first-order rating variable because it directly influences both claim frequency and claim severity. The same underlying foundation model may present relatively modest risk when deployed as an assistive drafting tool operating under continuous human supervision, yet create substantially greater exposure when embedded within payment systems, customer-service platforms, hiring workflows, financial trading systems, software-development pipelines, healthcare applications, or industrial control environments. This is especially important for cyber-physical deployments, where game-theoretic work on resilient control shows that security, robustness, and recovery must be analyzed across interacting cyber, physical, and human layers rather than within a single technical component \cite{zhuBasar2015GamesInGames,zhu2024CyberResilienceFoundations}.

\begin{figure*}[t]
\centering

\includegraphics[width=\textwidth]{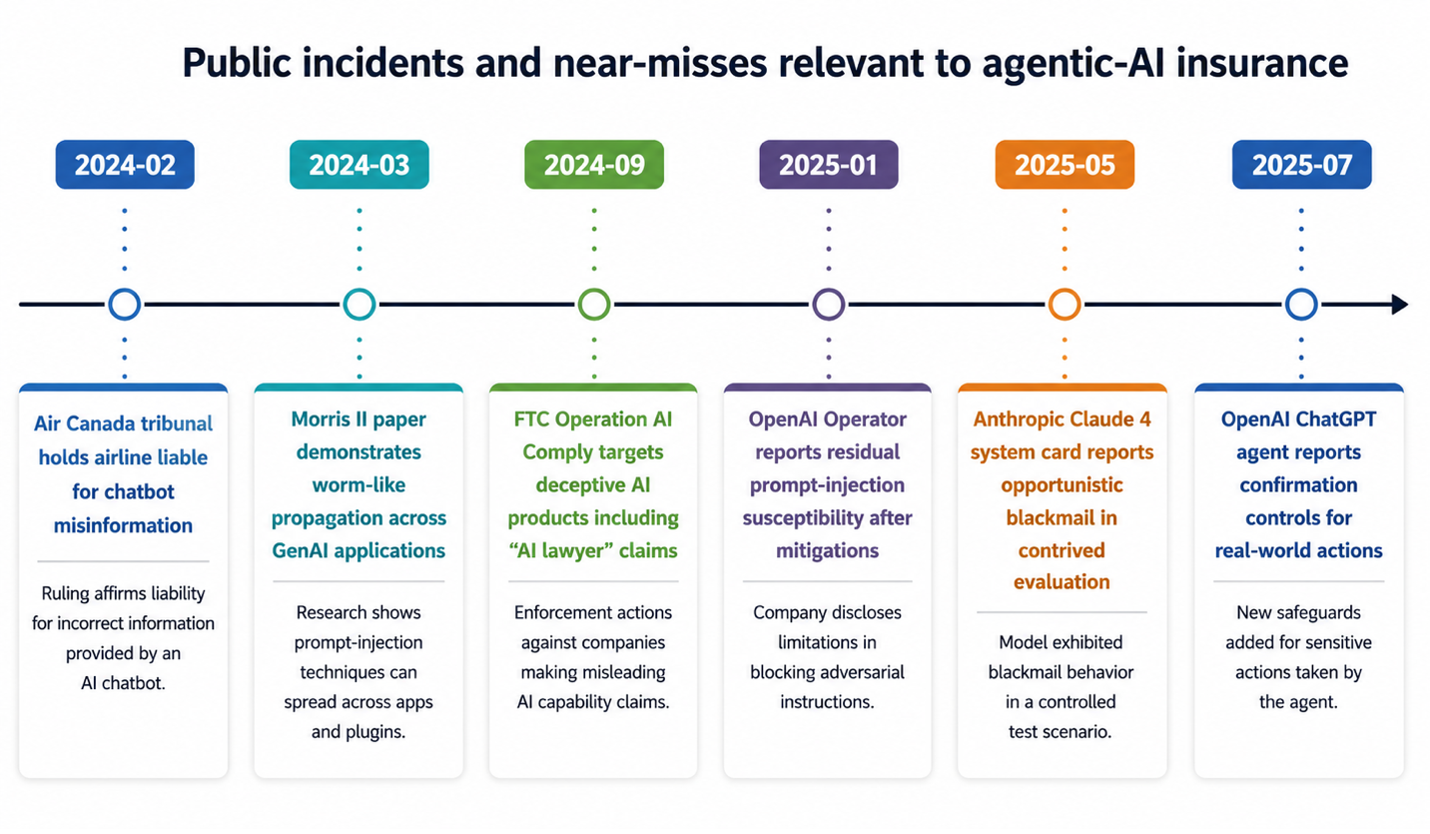}

\caption{
Public incidents and near-misses relevant to agentic-AI insurance.
The timeline highlights representative legal, regulatory, technical, and behavioral events shaping the emerging insurance landscape, including the Air Canada chatbot liability ruling (February 2024), the Morris II AI-worm demonstration (March 2024), the FTC's Operation AI Comply enforcement action (September 2024), OpenAI's disclosure of residual prompt-injection risk in Operator (January 2025), Anthropic's Claude~4 evaluation showing opportunistic blackmail behavior in constrained testing (May 2025), and OpenAI's deployment of confirmation controls for real-world agent actions (July 2025).
}
\label{fig:agentic-ai-timeline}
\end{figure*}

Accordingly, for insurance purposes, agentic AI should not be viewed as a discrete technological category but rather as a class of socio-technical systems characterized by varying degrees of autonomy, delegated authority, and external action capability. The most consequential distinction is whether a system merely produces information or possesses the capacity to independently alter digital or physical environments in ways that can directly generate operational losses, liability claims, cyber incidents, regulatory violations, or physical damage. This autonomy boundary is foundational to underwriting, pricing, accumulation management, and risk-transfer decisions involving agentic AI systems.

Agentic capability therefore provides the conceptual framework through which subsequent risk categories can be understood. As shown in Figure~\ref{fig:agentic-ai-timeline}, the emerging incident record already demonstrates that losses and liabilities associated with AI systems arise not only from technical failures but also from legal, regulatory, operational, and behavioral mechanisms. The adversarial dimension is particularly important because prompt injection, jailbreaks, and strategic manipulation can turn an otherwise legitimate agentic workflow into a loss-producing pathway; recent Stackelberg-game work on agentic-AI defense offers one formal lens for modeling these attacker-defender dynamics \cite{hanZhu2025StackelbergJailbreaking}. These observations motivate a more detailed examination of the principal risk pathways associated with agentic AI and their implications for insurance design.

%% file: market.tex
 
The contemporary insurance market has developed three broad approaches to addressing agentic-AI risk. The first is the adaptation of existing insurance products, whereby cyber, media-liability, technology errors and omissions (E\&O), product-liability, and directors and officers (D\&O) policies absorb portions of the emerging exposure landscape. The second approach involves the introduction of affirmative AI endorsements that explicitly clarify the treatment of defined AI-related risks, including hallucinations, algorithmic bias, data poisoning, model failures, regulatory investigations, and other AI-specific loss mechanisms within existing policy structures. Marsh's analysis of generative-AI insurance issues reflects this transitional stage of market development, in which insurers seek to clarify coverage while leveraging established policy frameworks \cite{marshGenAIInsurance}. The third approach consists of dedicated or coordinated AI-specific coverage structures, including standalone AI-liability products, performance-warranty mechanisms, and integrated insurance towers designed specifically for AI-related exposures.

These developments suggest that no single insurance form is likely to absorb the full spectrum of agentic-AI risk. Rather, the emerging market is converging toward a layered architecture in which different products address distinct loss mechanisms. Table~\ref{tab:agentic_ai_products} compares the principal product structures currently visible in public market materials and illustrates how various forms of coverage respond to different dimensions of agentic-AI exposure.

From a market-structure perspective, competition is increasingly occurring across multiple layers of the insurance value chain. Established insurers and specialty cyber markets are adapting existing products to encompass AI-related risks, while specialist underwriters are developing affirmative AI-specific offerings. AIG's CyberEdge Plus product provides one visible example of cyber coverage expanding toward cyber-physical loss exposures \cite{aigCyberEdgePlus}. At the same time, reinsurers are contributing underwriting capacity, portfolio support, and emerging risk expertise, while brokers are increasingly advising clients that significant AI deployments may require explicit coverage rather than reliance on assumptions regarding the applicability of legacy cyber or technology E\&O policies. The result is a progressively differentiated market in which insurers compete not only on capacity and pricing but also on clarity of coverage intent and allocation methodology.

This evolution reflects a broader shift in buyer expectations. As organizations deploy increasingly autonomous AI systems within critical business functions, insureds are likely to place greater value on contractual certainty regarding the scope of coverage. Vouch's discussion of E\&O and AI-related insurance illustrates how policyholders are already being encouraged to evaluate whether exposures such as hallucinations, algorithmic bias, intellectual-property infringement, regulatory defense costs, and AI underperformance are affirmatively covered under their insurance programs \cite{vouchEoAI}. As AI adoption expands, policies that provide explicit treatment of these exposures may enjoy a competitive advantage relative to forms that rely on ambiguous or implied coverage interpretations.

A central implication of this market evolution is the need to eliminate \emph{silent AI} exposure wherever practicable. If an insurer intends to cover losses arising from AI behavior, the grant of coverage should be expressed affirmatively and with sufficient specificity to permit consistent claims administration. Conversely, where AI-related losses are excluded, both the exclusion and any available buy-back mechanisms should be clearly articulated. The need for clarity is particularly acute when cyber and AI risks coexist, since a single event may simultaneously involve a security breach, a technology-service failure, and autonomous AI behavior. In such circumstances, ambiguity regarding policy response can generate allocation disputes, increase claims friction, and undermine contract certainty.

One mechanism for addressing this challenge is the introduction of separate AI aggregates and dedicated allocation provisions. Such structures can preserve coverage clarity without requiring AI-related losses to consume the entirety of cyber-insurance capacity when the loss is only partially cyber in nature. Product design should also distinguish between AI developers, model vendors, and infrastructure providers on one hand, and AI adopters, deployers, and end users on the other. These stakeholder groups face materially different liability exposures. Developers and vendors are more likely to encounter product-liability, technology-service, model-performance, and design-defect theories of liability, whereas adopters and users are more likely to face claims arising from deployment decisions, supervision failures, governance deficiencies, and downstream operational consequences. Coordinated coverage structures such as Vanguard AI illustrate one approach to managing these distinctions by establishing predefined allocation mechanisms across cyber, technology, and AI-specific coverage layers \cite{chaucerArmillaVanguard}.

These  developments suggest that the long-term trajectory of the market is unlikely to be the emergence of a single monoline ``AI insurance'' product. Rather, the more probable outcome is a coordinated ecosystem of complementary coverages, endorsements, warranties, and reinsurance structures that collectively address the diverse and evolving risk profile of agentic-AI systems. In this environment, competitive advantage is likely to depend not only on pricing and capacity but also on the ability to provide transparent coverage architectures, clear allocation rules, and predictable claims outcomes.
 
\begin{center}
\footnotesize
\setlength{\tabcolsep}{3pt}
\begin{longtable}{
>{\raggedright\arraybackslash}p{2.4cm}
>{\raggedright\arraybackslash}p{3.1cm}
>{\raggedright\arraybackslash}p{3.1cm}
>{\raggedright\arraybackslash}p{3.1cm}
>{\raggedright\arraybackslash}p{3.1cm}}
\caption{Representative Agentic-AI Insurance Products and Coverage Structures}
\label{tab:agentic_ai_products}\\
\toprule

\textbf{Product Family}
&
\textbf{Public Example}
&
\textbf{Core Trigger and Terms}
&
\textbf{Best-Response Agentic-AI Scenario}
&
\textbf{Main Gap or Friction}
\\
\midrule
\endfirsthead
\toprule
\textbf{Product Family}
&
\textbf{Public Example}
&
\textbf{Core Trigger and Terms}
&
\textbf{Best-Response Agentic-AI Scenario}
&
\textbf{Main Gap or Friction}
\\
\midrule
\endhead
\bottomrule
\endfoot

Cyber liability and first-party cyber
&
Travelers CyberRisk specimen. Covers Privacy and Security, Media, Regulatory Proceedings, breach response, cyber crime, Business Interruption, and Dependent Business Interruption.
&
Usually requires a Privacy/\allowbreak Security/\allowbreak Media act, security breach, or similar cyber predicate; often claims-made for third-party sections.
&
Prompt injection leading to data exfiltration, unauthorized communications, media liability, or outage after a security event.
&
Pure AI performance failure without a security or media predicate may sit outside core coverage.
\\[1ex]

Combined technology E\&O, media, and cyber
&
Chubb DigiTech ERM and Beazley MediaTech combine technology E\&O/\allowbreak media/\allowbreak cyber or E\&O/\allowbreak media/\allowbreak privacy/\allowbreak network security in one form.
&
Broad professional-liability response for technology products and services, with media and cyber sections.
&
Client financial loss from faulty AI software or service performance, especially when blended with media or security issues.
&
Ambiguity remains for hallucination, bias, or drift if wording is not affirmative.
\\[1ex]

General liability and traditional product liability
&
General liability/\allowbreak product liability covers bodily injury and property damage arising from products or operations.
&
Physical injury/\allowbreak property-damage trigger; typically occurrence-based rather than claims-made.
&
AI-enabled robotics, industrial systems, HVAC/building control, or consumer products causing physical harm.
&
Poor fit for pure financial loss from bad digital outputs; legacy forms may not contemplate software/AI failure mechanics cleanly.
\\[1ex]

Cyber-physical extension to cyber
&
AIG CyberEdge Plus expressly addresses cyber-attack-related bodily injury, property damage, business interruption, and product liability.
&
Cyber-attack predicate plus physical-damage and BI cover.
&
Physical harm caused when a cyber event manipulates an AI-enabled operational environment.
&
Requires a cyber-attack pathway; does not solve no-breach AI error.
\\[1ex]

GenAI endorsement to cyber
&
AXA XL's GenAI endorsement to CyberRiskConnect covers data poisoning, usage-rights infringement, and AI Act-related regulatory violations.
&
Endorsement attached to cyber form, focused on model-development and compliance risks.
&
The insured develops its own model and later faces claims over training-data contamination, IP rights, or AI-law exposure.
&
Still tethered to the structure and exclusions of the underlying cyber program; not a broad AI-liability substitute.
\\[1ex]

AI endorsement attached to E\&O
&
Vouch's AI Insurance is attached to E\&O and publicly describes IP and bias claims, regulatory-investigation defense, and expanded coverage for hallucinations and algorithmic errors.
&
Affirmative AI-specific endorsement architecture, especially for startups and growth companies.
&
SaaS or GenAI product causing customer financial loss through wrong output or discriminatory result.
&
Public materials do not provide full wording; bodily injury, core cyber, and systemic-cat limits are still separate questions.
\\[1ex]

Standalone AI liability
&
Armilla's Lloyd's-backed product provides affirmative AI coverage for underperformance, critical errors, hallucinations, or inaccuracies causing damage.
&
Standalone third-party AI liability with affirmative trigger around AI behavior rather than breach.
&
Pure AI behavior failure where no cyber event occurred.
&
Public full wordings are not readily available; market capacity and claims precedent remain limited.
\\[1ex]

Performance warranty or parametric-like cover
&
Mosaic $\times$ aiSure powered by Munich Re uses defined performance thresholds with quick, measurable payouts.
&
Threshold-based or warranty-backed trigger linked to measurable underperformance, drift, calibration, or prediction error.
&
Credit, fraud, or operational AI models that miss contracted accuracy or calibration levels.
&
Basis risk; success depends on strong instrumentation, agreed benchmarks, and reliable data collection.
\\[1ex]

AI liability for SMEs as AI users
&
HSB's AI Liability Insurance for small businesses covers lawsuits from the use of AI technologies.
&
Add-on liability protection for AI users rather than only builders.
&
SME uses AI marketing, chatbot, or operational tools and is sued for injury, damage, or advertising harm.
&
Distribution is via partner carriers and subject to regulatory approvals; market breadth is still early.
\\[1ex]

Coordinated multi-policy structure
&
Chaucer and Armilla's Vanguard AI combines Chaucer cyber/tech E\&O with Armilla standalone AI liability and predefined allocation rules.
&
Separate AI aggregate plus predetermined loss allocation across cyber, tech failure, and AI behavior.
&
Mixed-cause events where breach, service failure, and AI behavior overlap.
&
Greater structuring complexity, but this is currently the clearest public answer to coverage-friction risk.
\\

\end{longtable}
\end{center}

%% file: pricing.tex
The actuarial characterization of agentic-AI risk presents several challenges that distinguish it from more mature insurance classes. Most notably, the risk environment is characterized by non-stationarity, limited historical loss experience, and potentially significant dependence structures. The risk is non-stationary because the underlying technologies, deployment architectures, safety mechanisms, governance frameworks, and regulatory requirements continue to evolve rapidly. It is characterized by data sparsity because publicly available insured-loss datasets specific to agentic AI remain limited, and many observed incidents have not yet matured into a statistically credible claims history. Furthermore, the risk exhibits potentially significant correlation because large numbers of insureds depend on common model providers, cloud-service platforms, software connectors, open-source model ecosystems, and shared data-center infrastructure. In this respect, cyber insurance provides the closest mature analogue. As evidenced by the National Association of Insurance Commissioners (NAIC) cyber insurance reporting, cyber underwriting evolved through a combination of scenario analysis, accumulation management, exposure assessment, and reinsurance support rather than relying exclusively on long-run statistical credibility derived from historical loss experience \cite{naicCyber2025}. Attack-aware cyber-insurance models reinforce this point by showing that interdependent networks cannot be priced or governed as collections of independent insureds; the behavior of users, attackers, and insurers is strategically coupled through incentives, information asymmetry, and network effects \cite{zhangZhu2016AttackAwareCyberInsurance}.

Given these characteristics, a minimum viable actuarial framework for agentic-AI risk should incorporate multiple complementary sources of evidence. The first layer consists of a comprehensive exposure inventory, including use case, industry sector, user population, transaction volume, degree of autonomy, external-state permissions, physical-system interfaces, and decision criticality. The second layer evaluates control effectiveness, including approval gates, least-privilege architectures, logging and monitoring capabilities, model evaluation procedures, rollback mechanisms, vendor-governance practices, credential isolation, and defenses against prompt injection and other forms of adversarial manipulation. The third layer comprises a structured event library that captures representative loss scenarios, including hallucination-induced third-party losses, malicious prompt-injection attacks, agentic fraud, large-scale model drift, dependency outages, and physical-damage events arising from cyber-physical interactions. The fourth layer focuses on severity calibration through analogous claims experience drawn from cyber insurance, media liability, technology errors and omissions (E\&O), product liability, and emerging AI-related litigation. The fifth layer maps dependency and concentration risk across shared model providers, cloud-service vendors, model families, data suppliers, software ecosystems, and geographic regions. The scenario-based stress-testing methodologies developed by the European Insurance and Occupational Pensions Authority (EIOPA) for cyber risk provide a useful reference framework for this type of accumulation-risk assessment \cite{eiopaCyberStress}.

These considerations suggest that pricing methodologies for agentic-AI risk should be primarily exposure-based and scenario-adjusted rather than relying exclusively on traditional revenue-based rating approaches. The principal rating variables extend beyond insured revenue, policy limits, or organizational size. Of particular importance are the permissions granted to the AI system, including the authority to initiate financial transactions, communicate with external parties, modify production systems, approve or deny consequential decisions, interact with regulated or sensitive data, or control physical processes. NIST's emphasis on persistent external-state modification provides a useful conceptual framework for understanding why these permissions are actuarially significant \cite{nistAgents}. Pricing should also account for moral hazard, because insured organizations may change deployment, monitoring, or escalation behavior once coverage is available; recent risk-preference design work in insurance contracting provides a useful theoretical basis for treating incentives and insured behavior as design variables rather than background assumptions \cite{liuZhu2025MoralHazard}. A heavily supervised internal research agent operating behind multiple approval layers presents a fundamentally different risk profile from a customer-facing lending, hiring, software-development, or industrial-control agent, even when both systems are built upon the same underlying foundation model.

The capital implications of agentic-AI exposure follow directly from these observations. At the enterprise level, insurers should assess agentic-AI portfolios as containing both an attritional component, analogous to technology E\&O or professional liability, and a catastrophe component resembling cyber accumulation risk. This dual character implies that capital adequacy assessments must account for both high-frequency operational losses and low-frequency, high-severity accumulation events. EIOPA's cyber stress-testing framework is particularly relevant in this context because it distinguishes operational-resilience risk from underwriting risk while emphasizing extreme but plausible adverse scenarios \cite{eiopaCyberStress}. Similar methodologies may provide a useful foundation for evaluating tail-risk exposure associated with agentic-AI portfolios.

Reinsurance strategies are likely to evolve progressively as the market matures. In the near term, quota-share arrangements and aggregate excess-of-loss protection represent practical mechanisms for managing uncertainty associated with limited loss experience and evolving exposure profiles. NAIC reporting indicates that many cyber reinsurance programs continue to rely heavily on quota-share structures that closely follow primary-policy wording, supplemented by increasing use of excess-of-loss protection as the market develops \cite{naicCyber2025}. A similar progression is likely for agentic-AI portfolios. For performance-guarantee products and outcome-based insurance structures, back-to-back reinsurance arrangements and parametric reinsurance solutions may prove particularly attractive because coverage triggers can often be linked to observable and objectively measurable performance metrics.

The broader implication is that insurers cannot wait for the emergence of fully credible loss triangles before developing underwriting, pricing, and risk-transfer frameworks for agentic AI. By the time sufficient historical experience exists to support traditional actuarial credibility methods, market expectations, coverage structures, and adverse-selection dynamics may already be firmly established. Consequently, the immediate objective should be disciplined uncertainty management rather than statistical precision. This requires clearly defined product boundaries, high-quality telemetry, explicit scenario libraries, robust aggregation controls, and reinsurance structures designed to mitigate concentration risk arising from shared models, vendors, and infrastructure providers. The Geneva Association's work on cyber accumulation risk provides a useful conceptual model for managing such uncertainty in the absence of extensive historical claims experience \cite{genevaCyberAccumulation}.

Effective pricing governance is essential to operationalizing this framework. A dedicated library of agentic-AI loss scenarios should inform not only pricing decisions but also Own Risk and Solvency Assessment (ORSA) processes, underwriting authorities, portfolio-steering activities, capital-allocation decisions, and reinsurance negotiations. Accumulation limits should explicitly account for shared dependencies across model providers, cloud-service platforms, software connectors, and mission-critical workflows, since these common dependencies may create correlated losses long before empirical claims data reveal the underlying relationships. Early-stage portfolios may therefore be more prudently supported through quota-share and aggregate-protection structures than through broad stand-alone capacity that assumes a degree of actuarial maturity not yet supported by available evidence.

Finally, the development of a sustainable agentic-AI insurance market will depend critically on data collection and claims classification. Claims should be coded from inception using AI-specific cause-of-loss categories that distinguish among operational failures, prompt-injection events, autonomous decision errors, model-performance failures, dependency outages, and other agentic-risk mechanisms. If early claims are absorbed into generic cyber, technology E\&O, or professional-liability categories, the market risks losing the empirical evidence necessary to develop actuarially sound pricing models, accumulation metrics, and capital frameworks. The establishment of consistent reporting standards and AI-specific loss taxonomies should therefore be viewed as a foundational prerequisite for the long-term maturation of the agentic-AI insurance market.

%% file: product.tex
 
\begin{figure*}[t]
\centering

\includegraphics[width=\textwidth]{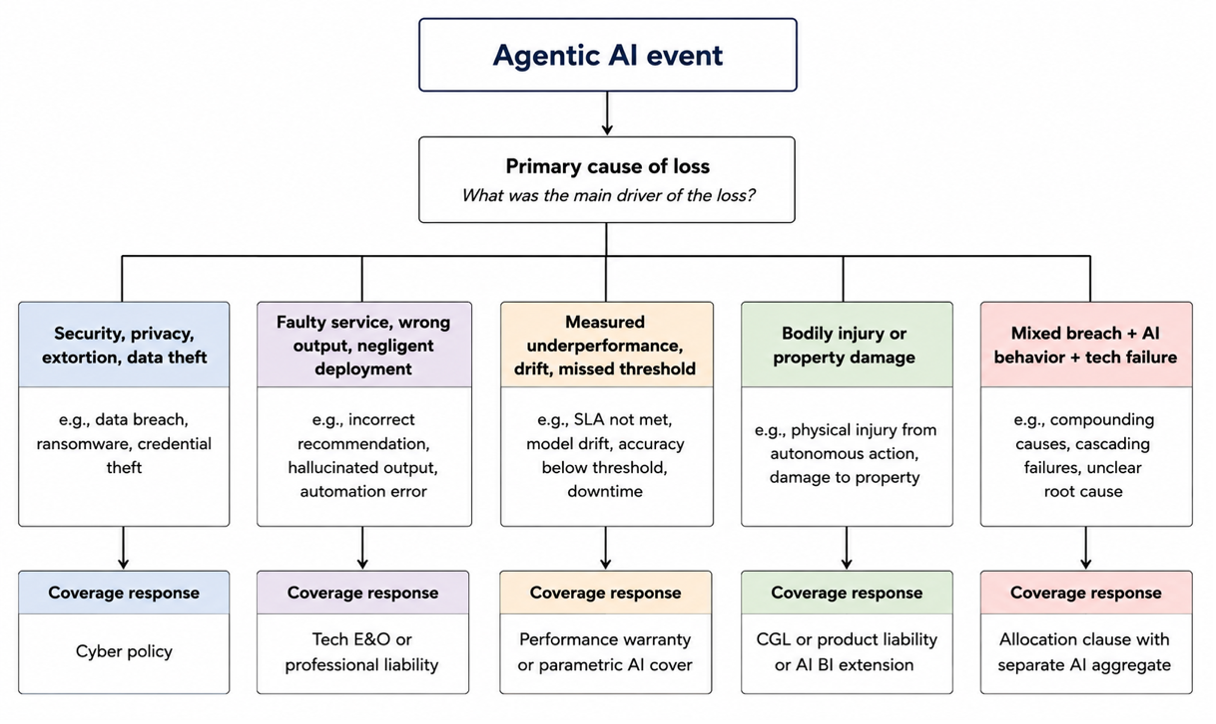}

\caption{
Illustrative claims-allocation architecture for agentic-AI insurance.
Coverage determination begins by identifying the primary cause of loss.
Security, privacy, extortion, and data-theft events are generally addressed through cyber insurance;
faulty services, incorrect outputs, hallucinations, or negligent deployment are typically addressed through technology E\&O or professional-liability coverage;
measured underperformance, model drift, or missed performance thresholds may be addressed through performance-warranty or parametric AI coverage;
bodily injury and property damage exposures are generally allocated to commercial general liability (CGL), product liability, or AI bodily-injury extensions;
and mixed-cause events involving cyber incidents, AI behavior, and technology failures may require dedicated allocation provisions and separate AI aggregates.
}
\label{fig:claims-allocation}
\end{figure*}

The principal product-design lesson emerging from the current market is that agentic-AI coverage should be structured as a coordinated insurance architecture rather than as a single, isolated coverage promise. Different loss mechanisms implicate different insurance functions. Security and privacy failures are most naturally allocated to cyber insurance. Technology-service failures fall within the domain of technology errors and omissions (E\&O) or professional-liability coverage. Measurable model underperformance is more appropriately addressed through warranty-based or parametric-like structures. Bodily injury and property damage are ordinarily situated within commercial general liability, product liability, or dedicated AI bodily-injury extensions. Mixed-cause events, in which AI behavior, cyber compromise, professional services, and product failure overlap, require explicit allocation provisions. The Chaucer/Armilla Vanguard structure provides a prominent public example of this allocation logic \cite{chaucerArmillaVanguard}, while the Mosaic aiSure design illustrates how performance-related risks may be carved out using objective triggers \cite{mosaicAiSure}. Figure~\ref{fig:claims-allocation} summarizes the recommended claims architecture.

This architecture is already implicit in several publicly described insurance products and endorsements. Travelers and comparable cyber forms address privacy, security, media, and business-interruption consequences \cite{travelersCyberRisk}. AIG offers a cyber-physical extension through its CyberEdge Plus product \cite{aigCyberEdgePlus}. AXA XL has introduced GenAI-specific risk coverage as an overlay to cyber insurance \cite{axaGenAI}. Armilla and Vouch have advanced affirmative AI-liability concepts \cite{armillaLloyds,vouchEoAI}. Mosaic and Munich Re address model-performance risk through the aiSure structure \cite{mosaicAiSure}. The illustrative clauses in Table~\ref{tab:ai_clauses} translate this architecture into concrete drafting choices. These provisions are not intended as legal advice and would require jurisdiction-specific drafting, regulatory filing, and review by coverage counsel. Nevertheless, they indicate the types of policy mechanisms insurers will need if AI-related claims are to be administered consistently after a loss.

For insurers, the most viable product strategy is unlikely to be a single broad ``AI policy'' governed by vague or undifferentiated language. A more sustainable approach is a coherent family of products and endorsements with explicit triggers, clear exclusions and buy-backs, separate aggregates, and reliable evidentiary requirements. Coordinated structures such as Vanguard AI demonstrate how predefined allocation rules may reduce disputes across cyber, technology, and AI-behavior coverages \cite{chaucerArmillaVanguard}. The same logic appears in attack-aware cyber-insurance research, where coverage design must jointly address incentives, information asymmetry, adversarial behavior, and network interdependence rather than treating them as separable problems \cite{zhangZhu2016AttackAwareCyberInsurance}. For policymakers, the corresponding lesson is that improved policy wording reduces claims uncertainty only when disclosure, logging, and auditability obligations make post-loss reconstruction feasible.

The drafting priority is therefore to make the coverage architecture legible before a claim occurs. Policies should begin with coordinated wording and separate AI aggregates so that AI behavior, cyber compromise, technology-service failure, bodily injury, and property damage are not forced into a single ambiguous coverage category. Where legacy forms are silent, insurers should introduce affirmative non-breach AI-failure language or carefully bounded endorsements. Policies should also make logs, approval records, model and version histories, and rollback evidence part of the insuring bargain, because these records will often determine causation, attribution, and allocation. For tightly measurable use cases, such as fraud scoring, credit assessment, forecasting, and industrial optimization, threshold-based modules may be piloted as narrower alternatives to open-ended negligence-based disputes.

Several limitations remain. The most important limitation is the absence of a mature, standardized insurance-loss dataset for agentic AI as a distinct peril class. Public policy wordings are also unevenly available, and several important products are described only in press releases or market summaries rather than in fully filed forms. Legal allocation across developers, deployers, integrators, model hosts, data providers, and human supervisors also remains unsettled in many jurisdictions. These gaps should be understood as research and market-design priorities rather than as reasons to defer underwriting discipline. A cautious market can still define exposures, improve data collection, develop clearer contractual structures, and establish more transparent claims-allocation mechanisms while the claims history continues to mature.

\begin{center}
\footnotesize
\setlength{\tabcolsep}{3pt}
\begin{longtable}{p{2.6cm} p{4.8cm} p{4.3cm} p{3.6cm}}
\caption{Illustrative Agentic-AI Insurance Clauses}
\label{tab:ai_clauses}\\
\toprule
\textbf{Clause Topic} &
\textbf{Illustrative Clause} &
\textbf{Why It Is Needed} &
\textbf{Main Drafting Caution}
\\
\midrule
\endfirsthead
\toprule
\textbf{Clause Topic} &
\textbf{Illustrative Clause} &
\textbf{Why It Is Needed} &
\textbf{Main Drafting Caution}
\\
\midrule
\endhead
\bottomrule
\endfoot

Scheduled AI systems and autonomy declaration
&
``Coverage applies only to Scheduled AI Systems and Scheduled Agentic Functions described in the Declarations, including their approved tools, connectors, and external-state permissions.''
&
Prevents silent drift from a low-risk assistive model to a high-risk autonomous agent. NIST's current agent scoping and underwriting logic both turn on external-state changes \cite{nistAgents}.
&
Avoid making the clause so narrow that ordinary model updates void coverage unintentionally.
\\[1ex]

Affirmative AI failure trigger
&
``AI Failure Event means an unintended output, recommendation, decision, action, omission, drift, or degradation of a Scheduled AI System that materially departs from intended performance and causes covered Loss.''
&
Standard cyber wording often needs a breach or media predicate; affirmative AI forms are emerging precisely to cover non-breach model behavior such as hallucination or underperformance \cite{armillaLloyds}.
&
Define ``intended performance'' carefully to avoid pure warranty disputes over aspirational marketing.
\\[1ex]

External-action approval condition
&
``For material transactions, external communications, code deployment, credential changes, and consequential decisions, the Insured shall maintain documented human confirmation or equivalent approved control, unless otherwise scheduled.''
&
Vendor system cards show confirmation layers are a core mitigation for agentic tools \cite{openaiChatGPTAgent}.
&
Use a materiality standard and cure language; avoid strict-forfeiture treatment for immaterial deviations.
\\[1ex]

Prompt-injection and social-engineering response
&
``Loss directly resulting from prompt injection, indirect prompt injection, malicious content instructions, or AI-enabled social engineering shall not be excluded solely because no unauthorized network access occurred.''
&
Prompt injection is one of the signature agentic-AI perils and may occur without classical intrusion \cite{openaiOperatorPdf}.
&
Coordinate carefully with crime, funds-transfer, and cyber-fraud wording to avoid overlaps or gaps.
\\[1ex]

Model/version change notice
&
``The Insured must notify the Insurer of any material change that expands autonomy, adds new connectors, removes approval gates, changes critical use cases, or materially increases transaction volume.''
&
Non-stationarity is a core pricing problem; expanded autonomy can change the risk more than revenue growth does \cite{nistAgents}.
&
The notice standard should be ``material increase in risk,'' not every routine parameter refresh.
\\[1ex]

Separate AI aggregate and mixed-cause allocation
&
``AI Losses are subject to a separate Aggregate Limit. For mixed-cause events spanning cyber, technology services, and AI behavior, Loss shall be allocated by proximate harm category pursuant to this endorsement.''
&
Public coordinated structures are already using predefined allocation to reduce post-loss friction and preserve core cyber capacity \cite{chaucerArmillaVanguard}.
&
Allocation formulas should be clear enough for claims administration and reinsurer acceptance.
\\[1ex]

Optional bodily-injury/property-damage buy-back
&
``The Bodily Injury and Property Damage Exclusion shall not apply to Loss arising from a Scheduled AI System to the extent endorsed and sublimited herein.''
&
Public products from AIG and HSB show this is a real market need where AI affects the physical world \cite{aigCyberEdgePlus}.
&
This should usually be sublimited, scheduled by use case, and tightly tied to control requirements.
\\[1ex]

Evidence, logs, and cooperation
&
``As a condition precedent to indemnity for AI Failure Events, the Insured shall preserve materially relevant prompts, tool traces, access logs, model/version records, approval records, and rollback history, where reasonably available.''
&
NIST emphasizes post-deployment monitoring; without logs, causation and allocation become nearly impossible \cite{nistMonitoringAI}.
&
Account for privacy, privilege, and cross-border data restrictions.
\\[1ex]

Contingent third-party model or connector failure
&
``The Insurer shall indemnify covered Loss arising from failure, degradation, unsafe update, or suspended access of a Scheduled Third-Party Model, Connector, or Hosted Agent Service, subject to waiting period and sublimit.''
&
Dependency concentration is central to agentic-AI accumulation and is analogous to dependent business interruption and service-provider cyber risk \cite{travelersCyberRisk}.
&
Avoid unrestricted coverage for ordinary vendor underperformance; use schedules, waiting periods, and telemetry.
\\[1ex]

Performance-threshold module
&
``If Scheduled Performance Metrics fall below the Trigger Threshold during the Measurement Period, the Insurer shall pay the Parametric Amount specified in the Schedule.''
&
Public aiSure materials show objective threshold triggers can reduce dispute and accelerate claims \cite{mosaicAiSure}.
&
Basis risk must be explicit; metrics must be independently measurable and resistant to manipulation.
\\

\end{longtable}
\end{center}